\documentclass{article}

\usepackage[final]{neurips_2026}

\usepackage[utf8]{inputenc} 
\usepackage[T1]{fontenc}    
\usepackage{inconsolata}

\usepackage{algpseudocode}  
\usepackage[ruled, linesnumbered, boxed]{algorithm2e}
\usepackage{parskip}
\usepackage{changepage}

\usepackage{color}
\usepackage{url}            
\usepackage{wrapfig}
\usepackage{caption}
\usepackage{cancel}
\usepackage{graphicx}
\usepackage{subcaption}
\usepackage{epsfig} 
\usepackage{times} 
\usepackage{booktabs}
\usepackage{booktabs}       
\usepackage{amsmath}
\usepackage{amsthm}
\usepackage{amssymb}  
\usepackage[table, dvipsnames, svgnames, x11names]{xcolor} 
\definecolor{ggdarkred}{HTML}{DC4437}
\definecolor{ggblue}{HTML}{4385F5}
\definecolor{gggray}{HTML}{B2B2B2}
\definecolor{gglightblue}{HTML}{A0C2FF}
\definecolor{ggorange}{HTML}{F5B400}
\definecolor{gggreen}{HTML}{109D59}

\usepackage{marvosym}
\definecolor{mygray}{gray}{.9}
\usepackage{pifont}
\usepackage{nicefrac}       
\usepackage{microtype}      
\usepackage[table, dvipsnames, svgnames, x11names]{xcolor}        
\usepackage{colortbl}

\usepackage{multirow}
\usepackage{cite}

\usepackage{enumitem}
\usepackage{graphics} 
\usepackage{svg}
\usepackage{bm}
\usepackage{float} 
\usepackage{stfloats}
\usepackage{adjustbox}
\usepackage{tablefootnote}
\usepackage[colorlinks,
linkcolor=blue,
anchorcolor=blue,
citecolor=blue,
urlcolor=Aqua,
backref=page]{hyperref}
\usepackage{makecell}

\usepackage{twemojis}
\usepackage{diagbox}

\newcolumntype{x}[1]{>{\centering\arraybackslash}m{#1pt}}

\title{ODEWorld: A Continuous Predictive Architecture\\ via Physical-Time Flow}
\author{
Dongxiu Liu$^{1*\dagger}$, 
Haoyi Niu$^{2*}\textsuperscript{\Letter}$, 
Peng Cheng$^{1}$, 
Yuan Gao$^{1}$, 
\textbf{Xirui Kang}$^{1\dagger}$, \\ 
\textbf{Sangli Teng}$^{2}$, 
\textbf{Koushil Sreenath}$^{2}$,
\textbf{Xianyuan Zhan}$^{1}\textsuperscript{\Letter}$\\
$^1$ Institute for AI Industry Research (AIR), Tsinghua University, Beijing, China\\
$^2$ Berkeley Artificial Intelligence Research (BAIR) Lab, University of California, Berkeley, USA\\
}

\begin{document}
	
	\maketitle
 {
    \renewcommand{\thefootnote}{\fnsymbol{footnote}}
    
    \footnotetext[1]{%
        Equal contribution, order determined by coin flip.  
        $\dagger$ Work done during internship at AIR.%
    }
    
    \addtocounter{footnote}{2} 
    \renewcommand{\thefootnote}{\textsuperscript{\Letter}}
    \footnotetext[3]{Corresponding authors: \texttt{niu@berkeley.edu}, \texttt{zhanxianyuan@air.tsinghua.edu.cn}.}
    
    \renewcommand{\thefootnote}{\arabic{footnote}}
}
     
	\begin{abstract}
In the physical world we inhabit, space and time are fundamentally continuous. However, existing machine learning paradigms for world modeling are largely confined to discrete-time prediction and inference, thereby exhibiting significant inefficiency in capturing the underlying dynamics of the physical world. To bridge this gap, we introduce Physical-Time Flow (\textbf{PT-Flow}), a novel approach that learns a continuous latent velocity field operating in physical time. Crucially, the underlying dynamics of sequential data are parameterized by an ordinary differential equation (ODE) embedded in a well-structured dynamical representation space. Under this paradigm, the prediction of the future can be recast as temporal integration via an ODE solver in the compressed latent space. Building upon PT-Flow, we construct \textbf{ODEWorld}, a continuous-time latent world model that is both efficient and versatile. By extracting time-variant features and enforcing ODE properties on both the dynamical representation space and the latent velocity field, ODEWorld effectively addresses the long-standing representation collapse issue in the latent world model literature. This also enables high-quality image reconstruction with ODEWorld even after long-horizon prediction. Moreover, its continuous nature allows for arbitrary temporal resolution and even backward prediction, which is impossible for most discrete-time models. Lastly, benefiting from the learned dynamics-centric latent space, ODEWorld can provide rich planning-oriented information to facilitate downstream policy learning. Comprehensive experiments across multiple simulation and real-world datasets demonstrate that ODEWorld successfully reconciles planning-conducive dynamics abstraction with visual realism, excelling in both video generation and robotic control.
\href{https://dstate.github.io/odeworld_website/}{Project Website}.
\end{abstract}
	
\section{Introduction}
The physical world we inhabit is fundamentally continuous in both space and time. In our daily life, objects move, interact, and collide in continuous time, governed entirely by underlying physical laws; meanwhile, biological organisms, such as humans and animals, are naturally attuned to the perception, understanding, and decision-making of continuous transitions. Despite this, contemporary AI models almost exclusively rely on discrete-time modeling paradigms, such as modeling video as sequences of image frames~\citep{ho2022video,bruce2024genie,Sora,he2026video,wang2026world} and solving decision-making tasks in discrete time steps~\citep{kim2026cosmos,sutton1998introduction,hansen2022temporal,li2025uva, ye2026world}. This inherent discrepancy makes modern predictive architectures highly inflexible when handling irregularly sampled data, imposing strict requirements on data for temporal alignment during both training and generation~\citep{ha2018world, hafner2023mastering}. Furthermore, simply relying on learning discrete-time pixel/sensor measurement transitions ignores the fundamental continuity in physics-grounded systems, causing current models to function primarily as ``video-mimicking'', and become inefficient in internalizing the intrinsic, higher-order physics inherent in reality~\citep{cellier2013continuous,krishnapriyan2023learning, cheng2024look, cheng2026sample}.

To bridge this gap, we introduce the concept of Physical-Time Flow (PT-Flow). At its core, PT-Flow conceptualizes the underlying dynamics in the data as a continuous velocity field in a well-structured latent space, where the latent state transition is smoothly governed by an ordinary differential equation (ODE) along actual physical time. To make such a continuous modeling framework possible, PT-Flow features two critical designs: 1) \textit{dynamical representation decoupling} and 2) \textit{direct first-order supervision}. The former decouples the time-varying, dynamics-related information from the data into a clean, ODE-compatible dynamical representation space, getting rid of the nuanced background and time-invariant details for efficient learning.
The latter directly supervises the latent velocity field with the approximated ground truth latent velocities through an elegant Jacobian-vector product (JVP) projection. The above two designs lead to a new predictive architecture, in which the prediction of the future can be recast as temporal integration via an ODE solver in the compressed latent space, in some sense similar to flow matching~\citep{lipmanflow}, but operating in physical time. 

Building upon PT-Flow, we construct a continuous-time latent world model, called \textbf{ODEWorld}, which embodies many desirable features. First, it disentangles dynamical representation by conditioning the encoder and decoder directly on static context, thereby offloading the burden of dynamical representation learning, allowing the latent space to focus exclusively on capturing essential dynamical changes. Second, ODEWorld circumvents the long-standing challenge of representation collapse in Joint Embedding Predictive Architecture (JEPA)-based frameworks~\citep{assran2023self, bardes2024vjepa, assran2025v}, where the latent space can degenerate into trivial solutions due to the indirect supervision of the coupled encoder and latent predictor. While some recent efforts~\citep{wang2026temporal, maes2026leworldmodel} have added external regularization schemes to mitigate this issue, however, at the expense of sacrificing model expressiveness.
By contrast, ODEWorld imposes direct supervision on the latent velocity field while harmoniously regularizing the ODE properties of the dynamical representations. These architectural choices empower ODEWorld to achieve high-fidelity image reconstruction with an extremely compact model, even across extended temporal horizons. Moreover, its continuous nature also enables arbitrary temporal resolution and even backward prediction--capabilities fundamentally absent in most discrete-time predictive models.

    Empirical evaluations across diverse simulation and real-world benchmark datasets, including LIBERO~\citep{liu2023libero} and Agibot-World~\citep{bu2025agibot}, demonstrate that ODEWorld effectively reconciles planning-conducive dynamics abstraction with visual realism. We show that the learned latent velocity field produces interpretable and smooth transformation paths between the given initial and goal states. This well-behaved continuous latent flow enables temporally consistent, long-horizon predictions by leveraging off-the-shelf ODE solvers within a highly compact latent manifold. Moreover, prediction by solving the continuous-time ODE naturally supports temporal super-resolution and robust video generation from irregularly sampled or dropped-frame data, where discrete-time models inherently fail. Lastly, robotic control experiments on the  LIBERO Benchmark also reveal that ODEWorld extracts rich, planning-centric representations that can greatly facilitate downstream policy learning. Ultimately, these capabilities establish ODEWorld as a highly versatile architecture, setting a new continuous recipe for both high-fidelity video synthesis and effective robotic policy learning.

	\section{Related Work}\label{rw}
    \paragraph{Predictive world models.} 
    Recent video generation models have achieved notable progress and shown great promise as scalable generative simulators of the physical world~\citep{Sora, yang2023learning, gen2act, bjorck2025gr00t, jang2025dreamgen}. These models learn to predict future observations as a surrogate for physical dynamics, producing full trajectory videos~\citep{du2023learning, robodreamer} or subgoal images~\citep{black2024zeroshot, cotvla}. These models have also garnered substantial attention from the robotics community, which translates their video prediction capability for control. A first line of methods adopts a planning-and-control paradigm, where actions are inferred from predicted videos, via inverse dynamics models~\citep{du2023learning, robodreamer}, goal-conditioned policies~\citep{black2024zeroshot, li2026causal, pai2025mimic,zheng2024instruction, li2024robo}, or direct pose extraction using off-the-shelf foundation models~\citep{liang2024dreamitate, chen2025large}. A second line of work instead integrates action generation into the world model, learning the joint distributions over future observations and actions~~\citep{li2025uva, ye2026world, kim2026cosmos, GR2, bi2025motus,  guo2024prediction, zhu2025unified}. However, all of these models suffer from architectural heaviness due to costly pixel-level reconstruction to derive world understanding, which have long been critiqued for their fundamental computational inefficiency.
    

    To address the efficiency issue, another branch of work focuses on learning latent world models, where predictions are performed in a compact representation space~\citep{assran2025v, maes2026leworldmodel, VPP,he2024learning, zheng2025universal,liang2025video, wang2025expressive, liu2025efficient,kang2026x}. These approaches construct latent representations through various objectives, including image reconstruction~\citep{VPP, xie2025latent}, contrastive learning~\citep{liu2025efficient,kang2025incorporating,nair2022r3m}, and explicit motion prediction~\citep{ATM, zheng2024tracevla, gao2024flip}, aiming to capture task-relevant structure while reducing the complexity of pixel-space modeling. An outstanding line of work is the JEPA family~\citep{assran2023self, wang2026temporal,maes2026leworldmodel, assran2025v,  balestriero2025lejepa}, which jointly learns to encode observations into a compact latent space and models transition dynamics by predicting future latents. While JEPA provides a compelling paradigm for latent world modeling, these methods can suffer from representation collapse, where the latent space degenerates toward trivial solutions, thus requiring carefully designed training strategies and regularization at the expense of expressivity loss~\citep{wang2026temporal, maes2026leworldmodel}. Both the aforementioned video world models and latent world models are exclusively constructed in discrete-time, which potentially become inefficient when capturing the fundamental higher-order physics in reality~\citep{cellier2013continuous,krishnapriyan2023learning}. In contrast, our proposed ODEWorld is directly modeled in continuous time while avoiding the representation collapse issue in JEPA frameworks by design, thereby offering great efficiency to model the physical world.

    \paragraph{ODE dynamics modeling.}
    Modeling continuous dynamical systems through ordinary differential equations has found extensive applications across diverse scientific domains, including physics~\citep{anderson1995computational, grigorenko2003chaotic,brunton2016discovering}, climate science~\citep{trenberth1992climate}, and neuroscience~\citep{izhikevich2007dynamical}. One prominent methodology involves explicitly constructing physics-informed learnable ODE dynamics~\citep{brunton2016discovering, champion2019data,greydanus2019hamiltonian,cheng2024look, cheng2026sample}. While these methods effectively embed known invariant properties into the network structure, they strictly require explicit prior knowledge of the true underlying governing equations or a given set of analytical terms to model the system dynamics, thereby severely limiting their expressiveness. Within this data-driven paradigm, classic approaches like Koopman operator theory~\citep{Mezi2005SpectralPO} are frequently employed in the control community. However, it also suffers from computationally expensive coordinate transformations and remains fundamentally restricted to an approximation of the true system dynamics. To circumvent the reliance on prior knowledge, neural ODE networks that learn neural ODE dynamic models directly from observation data have received increasing interest~\citep{chen2018neural, dupont2019augmented, liu2019neural, huh2020time}. Despite being promising, training such neural ODE networks often relies on computationally expensive adjoint methods~\citep{chen2018neural}, which are hard to scale and often suffer from instability during training. By contrast, in our proposed PT-Flow paradigm, we directly supervise the latent ODE velocity field with the approximated ground truth latent velocities, which reconciles the preservation of ODE properties with efficient and scalable training.

\section{Physical-Time Flow}\label{sec:pt_flow}
To go beyond the discrete-time modeling to a new architecture capturing continuous-time physical reality, we introduce the concept of \textbf{Physical-Time Flow (PT-Flow)}. The core idea of PT-Flow is to capture the underlying dynamics of high-dimensional data as a continuous velocity field within a proper latent space. 
Let $s_t \in \mathbb{R}^{d_s}$ denote the state input at physical time $t$, and $z_t \in \mathbb{R}^{d_z}$ be a properly encoded latent representation of $s_t$. Here, $t$ represents the physical time elapsed after starting at the initial state $s_0$. In our setting, $s_0$ can be flexibly selected from any frame in a demonstration rather than only the first frame. Accordingly, $t$ is defined as the relative temporal offset between $s_0$ and $s_t$. To capture the time-evolving dynamics of the physical system, a straightforward approach is to learn a time-dependent velocity field defined via the ordinary differential equation:
\begin{equation}\label{eq:ode}
\frac{dz_t}{dt} = v_\theta(z_t, t; z_0, c) \implies z_T = z_0 + \int_{0}^{T} v_\theta(z_t, t; z_0, c) dt
\end{equation}
where $z_0$ is the latent representation of the initial state $s_0$ and $c$ is the target conditioning (e.g., a goal state or a language instruction), which jointly specify the start and end state of the system.
If we can learn such a latent velocity field $v_\theta$, then future prediction can be redefined as a temporal integration task over the continuous physical time, thereby unlocking many interesting capabilities, such as arbitrary temporal resolution or backward prediction, as well as being amenable to many nice properties in control theory~\citep{cellier2013continuous}. The remaining questions are, what is the "proper" latent space and how can we learn $v_\theta$ with discrete-time supervision data?




\paragraph{Dynamical representation decoupling.} Obviously, not all dynamical systems can be modeled as an ODE function, especially for nuanced mixtures of visual features in a video. Ideally, for video data, we do not want to learn a velocity field to fit every static background detail and time-invariant object, but want to extract clean, essential dynamics-related representations $z_t$ out of the state space. To achieve this, we learn a pair of initial state-conditioned encoder and decoder: $f_{\text{dyn}}(s_t; s_0)=z_t$ and $g_{\text{dyn}}(z_t; s_0)=s_t$. By explicitly isolating static context (time-invariant part between $s_0$ and $s_t$) in both encoder and decoder, we make the encoded latent representation $z_t$ focus exclusively on dynamical changes. To prevent information loss, we introduce the following reconstruction loss:
\begin{equation}\label{eq:recon}
\mathcal{L}_{\text{dyn-recon}} = \mathbb{E}_{s_0,s_t\sim \mathcal{D}}\| g_{\text{dyn}}(f_{\text{dyn}}(s_t; s_0); s_0)-s_t \|^2.
\end{equation}
To further encourage the encoder $f_{\text{dyn}}$ to extract fundamental, ODE-compatible dynamical representations, we also enforce $f_{\text{dyn}}$ to also satisfy the ODE property, i.e., $\dot{z}_t=\frac{d f_{\text{dyn}}(s_t; s_0)}{d t}$. Next, we will show how this can facilitate the latent velocity field learning while achieving a direct, harmonious supervision on both $f_{\text{dyn}}$ and $v_{\theta}$.


\paragraph{Direct first-order supervision.} Note that the time derivative of the latent state satisfies
\begin{equation}\label{eq:jvp_proj}
    \dot{z}_t=\frac{dz_t}{dt}=\frac{d f_{\text{dyn}}(s_t; s_0)}{d t}=\frac{\partial f_{\text{dyn}}(s_t; s_0)}{\partial s_t}\cdot \frac{d s_t}{d t} + \bcancel{\frac{\partial f_{\text{dyn}}(s_t; s_0)}{\partial s_0}\cdot \frac{d s_0}{d t}}=\frac{\partial f_{\text{dyn}}(s_t; s_0)}{\partial s_t}\cdot \dot{s}_t
\end{equation}
As $s_0$ is time-invariant, hence $\frac{d s_0}{d t}=0$ and the related term vanishes. The above equation suggests that $\dot{z}_t$ can be directly obtained utilizing a Jacobian-vector product (JVP) projection, i.e., $\dot{z}_t=\text{JVP}(f_{\text{dyn}}(s_t; s_0),\dot{s}_t)$. As $\dot{s}_t$ can be directly approximated from training samples, i.e., simply calculating as $\dot{s}_t=(s_{t+1}-s_t)/\Delta t$ ($\Delta t$ is the time interval between two consecutive states) or adopting more advanced higher-order estimation tricks as we will describe in the next section, the evaluation of $\dot{z}_t$ is practically feasible. This enables us to introduce the following learning objective for the latent velocity field $v_{\theta}$, which directly supervises it with the approximated ground-truth latent velocities $\dot{z}_t$:
\begin{equation}\label{eq:v_obj}
    \mathcal{L}_v = \mathbb{E}_{s_0,s_t\sim \mathcal{D}, t} ||v_\theta(z_t,t;z_0,c)-\texttt{sg}(\text{JVP}(f_{\text{dyn}}(s_t; s_0),\dot{s}_t))||^2,
\end{equation}
We use the stop-gradient operation $\texttt{sg}(\cdot)$ to stabilize the optimization of the latent velocity field, following common practices in representation learning~\citep{chen2021exploring}. Interestingly, we find that PT-Flow remains effective even without stop-gradient, as shown in App.~\ref{app:detach}. This suggests that the proposed objective itself can provide effective constraints for learning meaningful representations. Unlike existing JEPA-style latent world models~\citep{assran2023self, bardes2024vjepa, assran2025v} that supervise the latent predictor via the consistency loss between current and predicted embeddings, PT-Flow employs a direct first-order supervision strategy to specifically optimize the latent velocity field, effectively decoupling the dynamics learning from embedding reconstruction. Such a design greatly mitigates the risk of representation collapse~\citep{wang2026temporal, maes2026leworldmodel} in JEPA frameworks, and ensures the latent manifold remains expressive and dynamically meaningful. 

By jointly optimizing the loss objectives $\mathcal{L}_{\text{dyn-recon}}$ and $\mathcal{L}_v$, we can learn a well-structured latent space and velocity field, in which the future prediction at an arbitrary time can be achieved by integrating with an off-the-shelf ODE solver in the compact latent space.

\begin{figure}[t]
    \centering
    \includegraphics[width=\textwidth]{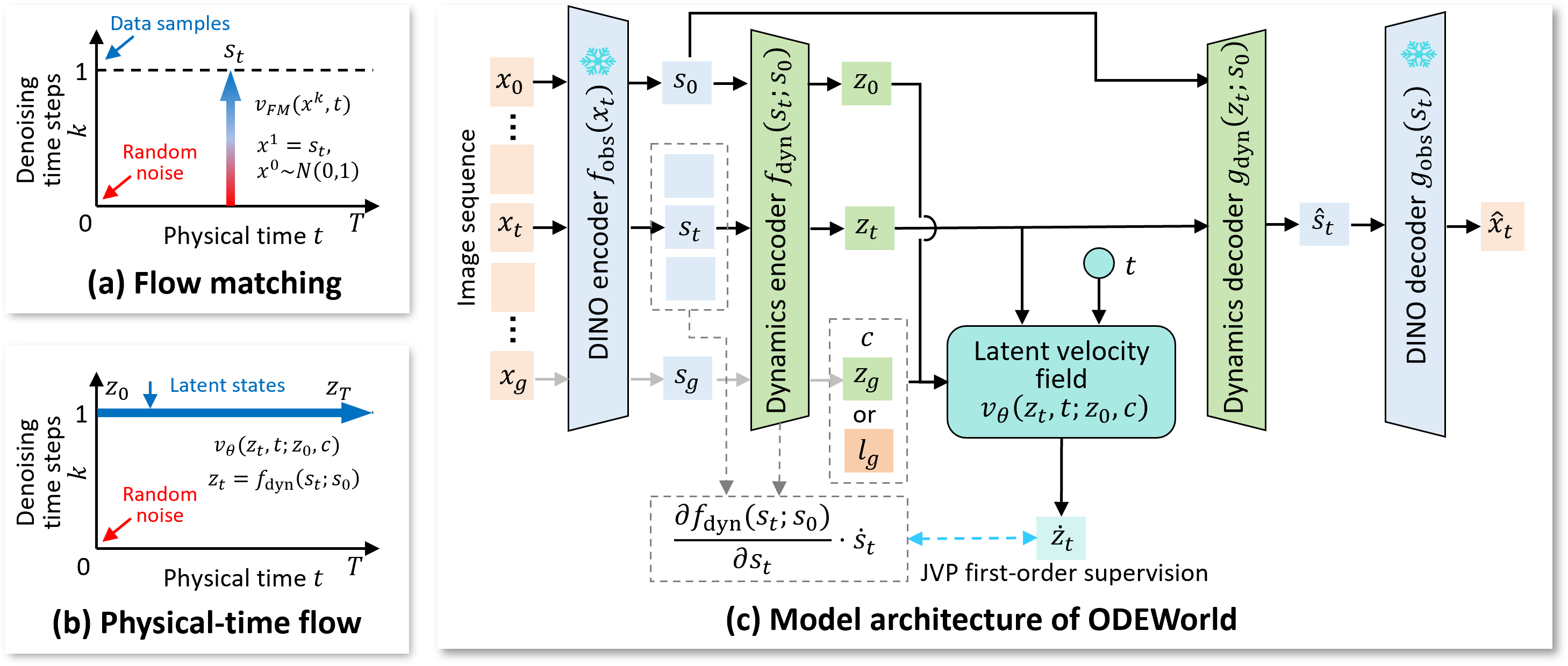}
    \caption{\small Illustration of PT-Flow and ODEWorld. (a), (b) Comparison between flow matching and our proposed PT-Flow. Although both methods learn time-dependent velocity fields and perform generation with ODE solvers, their notion of time and design philosophies are completely different. (c) Model architecture of ODEWorld.}
    \label{fig:framework}
\end{figure}

\section{ODEWorld: Latent World Modeling via PT-Flow}\label{sec:odeworld}
In this section, we instantiate the PT-Flow paradigm as \textbf{ODEWorld} (illustrated in Fig.~\ref{fig:framework}(c)), a highly versatile and efficient continuous-time predictive architecture for both high-fidelity video generation and effective robotic policy learning.


\textbf{Architecture overview.} 
Apart from building PT-Flow, we first employ a frozen, pre-trained DINOv2 encoder~\citep{oquab2024dinov2} as our vision backbone $f_{\text{obs}}$ to project high-dimensional raw image observations $x_t$ into a compressed feature space $s_t$. This transformed state space offers superior numerical stability compared to raw pixels when estimating the velocity target $\dot{s}_t$ in Eq. (\ref{eq:jvp_proj}).
Correspondingly, we train a dedicated image decoder $g_{\text{obs}}$ to reconstruct the DINO feature back to image space $\hat{x}_t$ for video generation. We follow the same recipe with RAE~\citep{zheng2026diffusion} to train $g_{\text{obs}}$ with a combination of L1, LPIPS~\citep{LPIPS} and adversarial losses~\citep{GAN}:
\begin{equation}\label{eq:rec_loss}
\begin{aligned}
s_t &= f_{obs}(x_t), \quad \hat{x}_t = g_{obs}(s_t), \quad x_t\in \mathcal{D}\\
\mathcal{L}_{\text{DINO-recon}}(x_t)
&= \mathcal{L}_{1}(\hat{x}_t, x_t)
+ \omega_L \mathcal{L}_{\text{LPIPS}}(\hat{x}_t, x_t)
+ \omega_G \lambda \mathcal{L}_{\text{GAN}}(\hat{x}_t, x_t).
\end{aligned}
\end{equation}


We further embed PT-Flow within the DINO feature space, where DINO image encoder and decoder $f_{obs}$ and $g_{obs}$ are frozen during the training of PT-Flow. We find that PT-Flow is sufficiently expressive to maintain consistency in latent planning and reconstruction within the DINO feature space, ultimately enabling high-fidelity video generation through $g_{\text{obs}}$.

Following the design of PT-Flow, we introduce an initial state-conditioned dynamics encoder $f_{\text{dyn}}$ and a decoder $g_{\text{dyn}}$, both of which are implemented using cross-attention blocks. The dynamics encoder $f_{\text{dyn}}(s_t; s_0)=z_t$ extracts the dynamical representation $z_t$ from the current state $s_t$ conditioned on the initial state $s_0$, where a few learnable query tokens attend to both $s_t$ and $s_0$ as keys and values, producing a compact latent representation that summarizes temporal changes between the initial and current states.
The decoder $g_{\text{dyn}}$ shares a similar structure, again with the initial state $s_0$ injected but serves as the query, while the latent representation $z_{\text{t}}$ acts as keys and values. This design forces the model to reconstruct $s_t$ by combining static context from $s_0$ with dynamical information encoded in $z_t$, thereby preventing $z_t$ from carrying static content. 

In our experiments, we find that a single token is sufficient to model $z_t$ while achieving strong performance. Specifically, $z_t \in \mathbb{R}^{1 \times 768}$ is highly compact compared to $s_t \in \mathbb{R}^{16 \times 16 \times 768}$ in original DINO space. More importantly, this result further supports the key insight of latent world modeling~\citep{assran2023self, maes2026leworldmodel}: raw pixels contain substantial redundancy, while the underlying dynamics can be effectively captured by a much smaller latent representation.
Based on the compact $z_t$, we parameterize the velocity network $v_\theta(z_t,t;z_0,c)$ by a lightweight 3-layer MLP. The time $t$ is incorporated by Feature-wise Linear Modulation (FiLM) layers~\citep{perez2018film}, while other parts $z_t,z_0,c$ are concatenated and fed into the model as input. Such a compact latent dynamics model design makes the latent planning and prediction with the ODE solver extremely efficient, while maximally preserving information for image reconstruction.


\textbf{Additional enhancements on PT-Flow.} 
 To ensure numerical stability, instead of modeling the latent velocity $v_{\theta}(z_t,t;z_0,c)$ with the actual physical time $t$ as in Eq. (\ref{eq:ode}), we implement $v_{\theta}(z_t,\tau;z_0,c)$ based on a rescaled physical time $\tau$, where $\tau = \frac{t}{L}$ and $L$ is a normalizing horizon length which is approximately the desired planning length. 
 This will make sure $\tau$ mostly resides within the range $[0, 1]$ during training.
 Under this reparameterization, $\dot{s}_\tau$ corresponds to $\dot{s}_t$ scaled by $L$ in time dimension. We assume that the system reaches a stationary state once the target condition $c$ is reached (e.g., the goal state $z_g$ or language instruction $l_g$). Therefore, when $\tau$ exceeds the rescaled time index $\tau_c$ of the last frame in a training sequence ending with condition $c$, we fix the future state $s_{\tau>\tau_c}$ to the last frame and drive the velocity field $v_\theta$ to zero in the end.

To enhance the estimation quality of the target velocity $\dot{s}_t$ for JVP projection in Eq. (\ref{eq:v_obj}), we employ a Savitzky–Golay derivative filter~\citep{saramaki1993finite}, which performs local polynomial regression over a sliding window. Specifically, given a sequence of states $\{s_{t-k}, \dots, s_{t+k}\}$ within a window of size $2k+1$, the target velocity is calculated as a linear convolution: $\dot{s}_t = \sum_{i=-k}^{k} w_i s_{t+i}$,
where $w_i$ are fixed coefficients obtained by fitting a low-order polynomial in a least-squares sense. In our implementation, we instantiate this filter with a window size of $2k+1=5$, yielding a filter kernel $w = \frac{1}{10}[-2, -1, 0, 1, 2]$. Unlike conventional smoothing kernels that blindly average signals (e.g., Gaussian blurring) or standard finite difference kernels, the chosen Savitzky–Golay kernel provides a decent compromise between numerical accuracy and temporal smoothness, due to its adoption of local least-squares polynomial fit. This ensures that the projected target $\dot{z}_t$ (via Eq.~(\ref{eq:jvp_proj})) is both temporally smooth and physically accurate, which substantially enhances the learning performance of the latent velocity field $v_{\theta}$.

\textbf{Training and inference pipeline.} With the pre-trained frozen DINO encoder $f_{\text{obs}}$ and decoder $g_{\text{obs}}$, the training of ODEWorld strictly follows the recipe of PT-Flow described in Section~\ref{sec:pt_flow}, which jointly minimizes the dynamics reconstruction loss $\mathcal{L}_{\text{dyn-recon}}$ (Eq.~\ref{eq:recon}) and the velocity supervision loss $\mathcal{L}_{v}$ (Eq.~\ref{eq:v_obj}), i.e., 
\begin{equation}\label{eq:overall_obj}
\mathcal{L} = \lambda_{rec}\mathcal{L}_{\text{dyn-recon}} + \lambda_v\mathcal{L}_{v}
\end{equation}
In our experiments, we find that simply setting the weighting parameters $\lambda_{rec}=\lambda_v=1$ without any tuning generally leads to reasonably good results. This further demonstrates the learning robustness of our proposed PT-Flow framework.

During inference, ODEWorld serves as a continuous-time ODE engine for both high-fidelity video synthesis and efficient latent planning. 
The inference pipeline works as follows:
given an initial raw observation $x_0$ and a target condition $c$, the model first projects the observation into the DINO space $s_0 = f_{obs}(x_0)$ and then encodes this state into the dynamical latent space $z_0 = f_{dyn}(s_0; s_0)$.
We can employ any off-the-shelf ODE solver, e.g. fourth-order Runge-Kutta method (RK4), to integrate the learned velocity field $v_\theta$ by $z_\tau = z_0 + \int_{0}^{\tau} v_\theta(z_t, t; z_0, c) dt$ to predict the latent state $z_\tau$ at arbitrary time $\tau \in [0, 1]$.
The evolved latent state $z_\tau$ can then be used for: 1) future video frame generation: the latent state is sequentially mapped back to the pixel space with $\hat{s}_\tau = g_{dyn}(z_\tau; s_0)$ and $\hat{x}_\tau = g_{obs}(\hat{s}_\tau)$; and 2) downstream policy learning: $z_\tau$ serves as a compact, dynamically meaningful subgoal. Its low-dimensional dynamics-centric nature provides a stable target for policy learning, bypassing the complexity of high-dimensional visual planning.
 A notable advantage of ODEWorld lies in its remarkable inference speed. By operating within the compact latent manifold $z$ and utilizing a shallow MLP to parameterize the velocity field $v_\theta$, the numerical integration becomes computationally lightweight. This efficiency allows for real-time video generation and fast iterative latent planning, facilitating responsive closed-loop control in complex environments.

\section{Experiments}\label{exp}


\subsection{Experimental Setups}
\textbf{Datasets. }
We train ODEWorld on LIBERO~\citep{liu2023libero} simulation dataset and AgiBot-World~\citep{bu2025agibot} real-world robot datasets. LIBERO consists of 130 tasks with 6,500 high-quality human teleoperation demonstrations, covering diverse motions and atomic skills. We use the full dataset to train ODEWorld and all baseline models for image generation. For the AgiBot-World, we train ODEWorld on the AgiBotWorldChallenge-2025 subset, a large-scale embodied manipulation benchmark built upon the AgiBot World Colosseo platform~\citep{bu2025agibot}. The dataset contains approximately 30K manipulation trajectories with an average length of around 260 frames per trajectory, providing diverse robot-object interaction behaviors with visual observations across various real-world manipulation tasks. More details about datasets are provided in App.~\ref{app:dataset_detail}.



\begin{figure}[t]
    \centering
    \includegraphics[width=\textwidth]{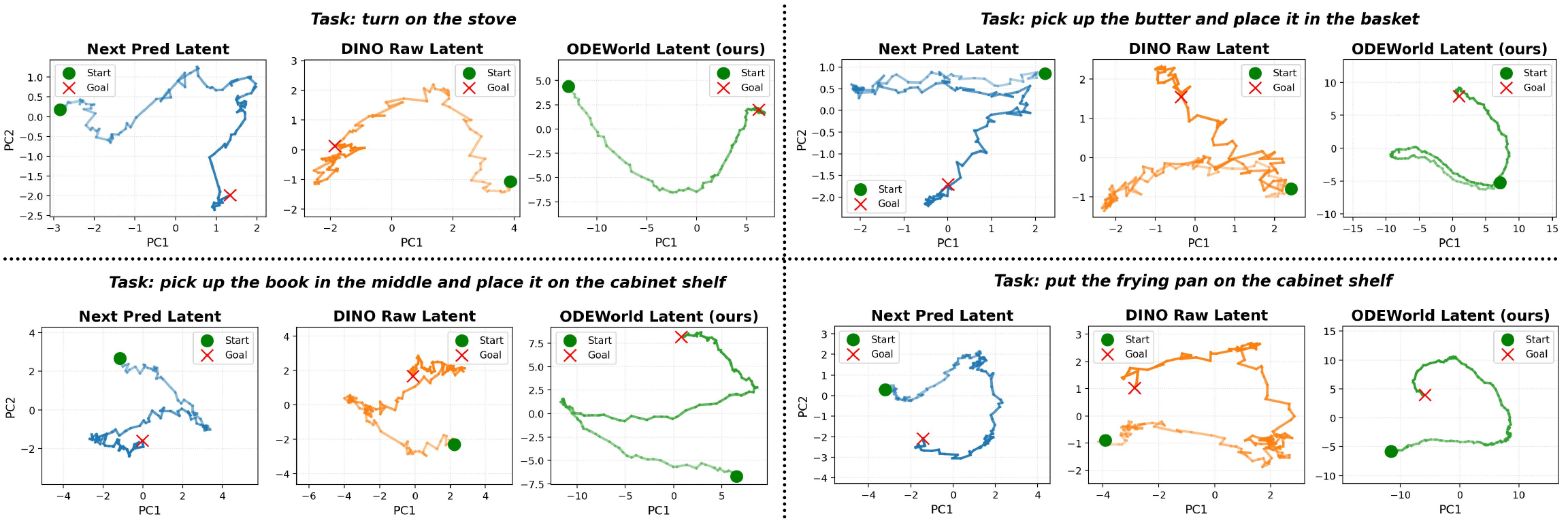}
    \caption{\small\textbf{Data trajectories in different latent spaces.} ODEWorld produces significantly smoother trajectories in its dynamical latent space across various LIBERO tasks compared to the latent spaces of DINO and the next-step prediction baseline.}
    \label{fig:latent_smooth}
\end{figure}

\textbf{Baselines. } 
To evaluate the video generation quality of ODEWorld, we compare it against latent world models that also support reconstruction from semantic representations. Specifically, we consider Latent Diffusion Planning (LDP)~\citep{xie2025latent} and V-JEPA 2~\citep{assran2025v}. LDP employs a reconstruction-based VAE latent space and performs planning and policy learning within this latent space, with generated latents decoded back into images via a VAE decoder. V-JEPA 2 jointly learns a representation encoder and a latent predictor, and similarly decodes predicted latents into raw images through a learned decoder. These models serve as suitable baselines for assessing generative quality in latent world model settings. As for policy learning, we compare ODEWorld against planning-based baselines, including the editing-based subgoal planner SuSIE~\citep{black2024zeroshot} and the end-to-end predictive inverse dynamics model Seer~\citep{tian2025predictive}, the video prediction policy~\citep{VPP}, and a vanilla goal-and-language-conditioned behavior cloning (GLCBC) policy. More details about these baselines are provided in App.~\ref{app:video_gen_baseline} and App.~\ref{app:policy_baseline}.

\subsection{Visualization}
\begin{figure}[t]
    \centering
    \includegraphics[width=0.85\textwidth]{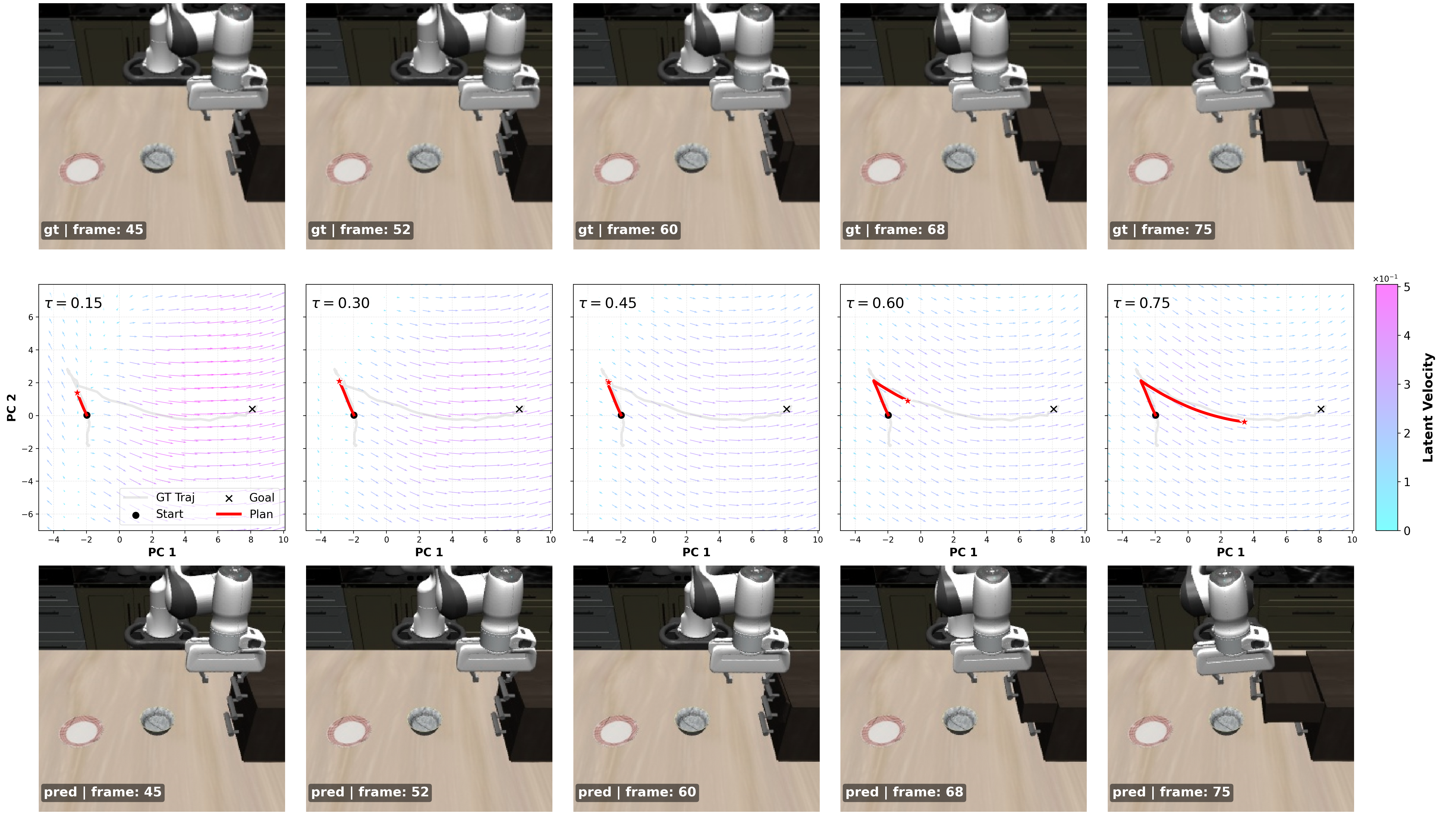}
    \caption{\small\textbf{Velocity field and latent planning.} The middle row shows visualization of globally consistent goal-directed convergence of velocity field and smooth latent plans from ODEWorld that adhere well to the ground truth. The top row shows ground truth frames, and the bottom row shows synthesized frames. }
    \label{fig:vel_field}
\end{figure}
\begin{figure}[t]
    \centering
    \includegraphics[width=0.87\textwidth]{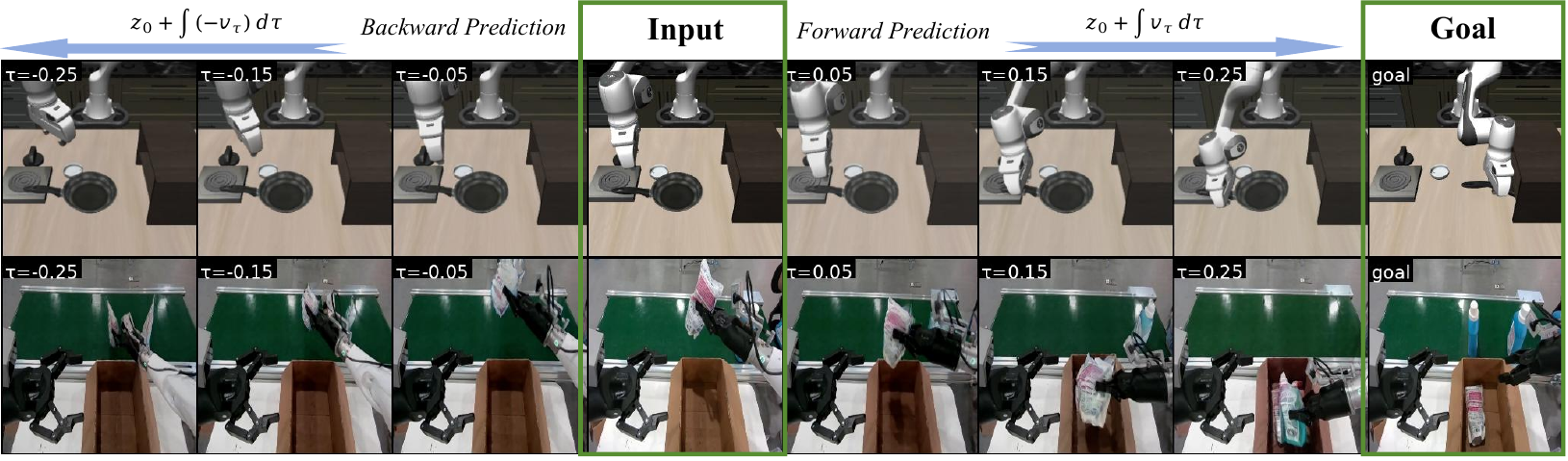}
    \caption{\small The continuous formulation enables ODEWorld to perform both forward and backward prediction, with motion consistent with real-world dynamics. We present the results on LIBERO benchmark and Agibot World. More results are provided in the Appendix.}
    \label{fig:for_back}
\end{figure}

\textbf{Latent space analysis. }
In Fig.~\ref{fig:latent_smooth}, we compare the latent space of ODEWorld against the raw DINO feature space and the latent of a discrete next-step predictive baseline. For the latter, we replace $v_\theta$ with a discrete transition model $z_{t+1}=p_\theta(z_t; c)$, remove dynamical representation decoupling designs in $f_{dyn}$ and $g_{dyn}$, and then optimize $p_\theta$ with JEPA-style objectives~\citep{assran2023self}. PCA visualizations for latent trajectories from multiple LIBERO tasks reveal that ODEWorld yields the smoothest latent trajectories. This is attributed to our direct first-order supervision, which regularizes the manifold by aligning it with an analytically filtered velocity target that effectively suppresses high-frequency jitters.
Crucially, while ODEWorld enhances dynamical smoothness, it maintains a manifold topology remarkably similar to the original DINO space. 
Unlike over-regularized models that tend to straighten the latent space to achieve smoothness at the expense of expressivity, ODEWorld's latent space preserves the principal components and the relative geometric structure in the original space. This structural fidelity ensures that the learned latent space retains rich semantic information, providing a high-fidelity foundation for downstream tasks such as video generation and subgoal planning for policy learning.

\textbf{Velocity field and open-loop latent planning. }
In the middle row of Fig.~\ref{fig:vel_field}, we sample a uniform grid in the projected PCA space and utilize the learned velocity field $v_\theta$ to compute the instantaneous latent velocities at each point. These are visualized as a vector field with arrows, the color of which represents the magnitude of dynamical transition. Starting from a latent state at a random time of the sampled trajectory, we perform a long-horizon rollout using an ODE solver (RK4) to integrate the velocity $v_\theta$ along normalized time $\tau$ without any further environmental observations.
To verify semantic fidelity, the integrated latent trajectory $z_\tau$ is passed through the decoders to synthesize image-space predictions shown in the bottom row, which are then compared against the ground-truth video sequence shown in the top row.
The visualization reveals three key properties:
(1) The velocity field captures the time-varying magnitude of the dynamics: in this example, the robot first approaches the handle and then pulls the drawer out, and the velocity correctly decays toward zero as the robot reaches the handle before accelerating again into the pulling phase.
This shows that the learned velocity field encodes not only the direction toward the goal, but also the correct instantaneous speed at each stage of the task.
(2) The model demonstrates high physical consistency even during open-loop planning. The planned trajectory (red line) adheres closely to the ground-truth latent path (gray line), while effectively filtering out the high-frequency noise and jitter present in the latter.
(3) Finally, the synthesized predictions in the bottom row remain spatially coherent and semantically accurate. 
This combination of planning accuracy and smoothness provides a stable, jerk-free foundation for downstream high-fidelity video generation and precise subgoal planning for policy learning. 

\begin{table}[t]
\centering
\caption{\small Comparison on video prediction quality and latency.}
\label{tab:video_baseline}
\begin{tabular}{x{85}x{35}x{35}x{45}x{35}x{35}x{45}}
\toprule
\multirow{2}{*}{Methods}
& \multicolumn{3}{c}{Short Fragment @16frames}
& \multicolumn{3}{c}{Long Fragment @64frames} \\
\cmidrule(lr){2-4} \cmidrule(lr){5-7}
& PSNR$\uparrow$ & LPIPS$\downarrow$ & Latency(s)$\downarrow$
& PSNR$\uparrow$ & LPIPS$\downarrow$ & Latency(s)$\downarrow$ \\
\midrule
LDP 
& 16.33 & 0.489 & 1.104 
& 16.27 & 0.461 & 3.953 \\

V-JEPA 2 
& 17.60 & 0.157 & 0.123 
& 16.47 & 0.166 & 0.619 \\

\rowcolor{gray!20}
ODEWorld (Ours) 
& \textbf{20.53} & \textbf{0.109} & \textbf{0.030}
& \textbf{19.46} & \textbf{0.134} & \textbf{0.072} \\
\bottomrule
\end{tabular}

\bigskip
\caption{\small On the LIBERO-LONG benchmark, ODEWorld achieves consistent improvements in policy performance. Averaged on 3 random seeds. Tasks are specified in App.~\ref{app:dataset_detail}.}
\label{tab:libero_long_exp}
\resizebox{\textwidth}{!}{
\begin{tabular}{x{150}x{20}x{20}x{20}x{20}x{20}x{20}x{20}x{20}x{20}x{20}x{50}}
\toprule
\resizebox{60pt}{!}{\diagbox{Method}{$\text{Task ID}$}} & 1 & 2 & 3 & 4 & 5 & 6 & 7 & 8 & 9 & 10 & Avg. Suc $\uparrow$ \\ \hline
GLCBC & 83.3 & 93.3 & 83.3 & 93.3 & 76.6 & 63.3 & 73.3 & 83.3 & 50.0 & 80.0 & 78.0 \\
SuSIE & 83.3 & 63.3 & \textbf{96.6} & \textbf{100.0} & 83.3 & 83.3 & 83.3 & 39.9 & 53.3 & 76.6 & 76.3 \\
Seer & 88.3 & 90.0 & 91.6 & 81.6 & 85.0 & 65.0 & \textbf{86.6} & 80.0 & 51.6 & 66.6 & 78.6 \\
VPP & \textbf{96.7} & \textbf{100.0} & 73.3 & 93.3 & 80.0 & \textbf{96.7} & 50.0 & 76.7 & \textbf{70.0} &73.3 & 81.0 \\
\rowcolor{gray!20}ODEWorld (Velocity) & 80.0 & 93.3 & 76.6 & \textbf{100.0} & 73.3 & 80.0 & 83.3 & 86.6 & 66.6 & 83.3 & 82.3 \\
\rowcolor{gray!20} ODEWorld (Single subgoal) & 86.6 & 86.6 & 86.6 & 96.6 & \textbf{86.6} & 86.6 & 80.0 & 83.3 & 46.6 & \textbf{86.6} & 82.6 \\
\rowcolor{gray!20} ODEWorld (Sequential subgoals) & 90.0 & 96.6 & 90.0 & \textbf{100.0} & 73.3 & 90.0 & 76.6 & \textbf{90.0} & 60.0 & 80.0 & \textbf{83.6} \\
\bottomrule
\end{tabular}
}

\bigskip
\caption{\small Real-world robot experiments with ODEWorld-guided policy learning.}
\label{tab:real_robot_odeworld}
\begin{tabular}{x{150}x{35}x{35}x{35}x{55}x{35}}
\toprule
Method & Shrimp & Mouse & Pen & Rearrangement & Average\\
\midrule
X-VLA & 80\% & 60\% & 30\% & 50\% & 55\%\\
\rowcolor{gray!20}
ODEWorld (Sequential subgoals) & \textbf{90\%} & \textbf{80\%} & \textbf{70\%} & \textbf{80\%} & \textbf{80\%}\\
\bottomrule
\end{tabular}
\end{table}

\begin{figure}[t]
    \centering
    \includegraphics[width=0.87\textwidth]{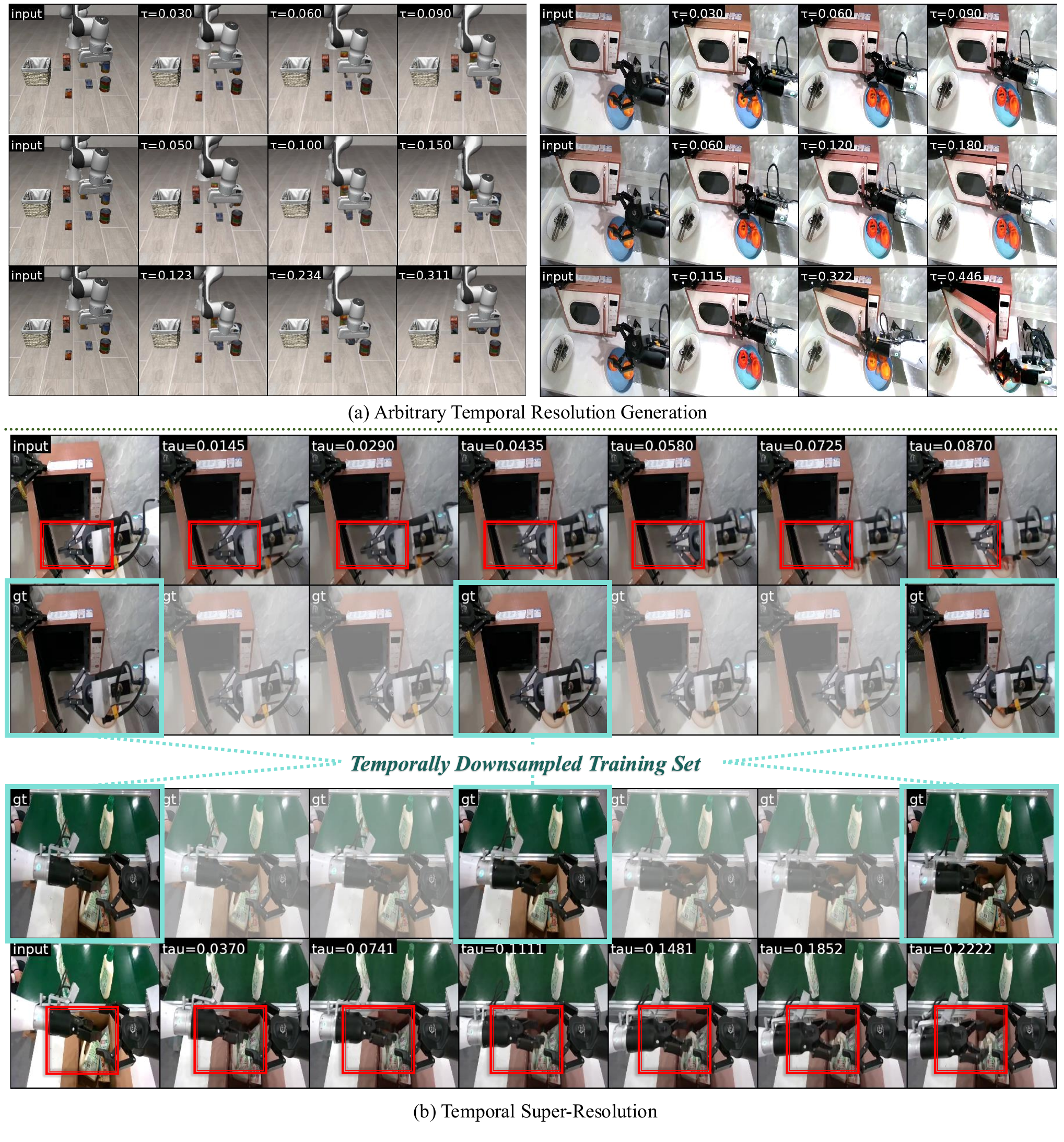}
    \caption{\small ODEWorld allows any resolution generation with flexible temporal speed. Based on this, ODEWorld can generalize beyond the sampling rate of the training data. In particular, when trained on temporally downsampled sequences, the model is able to generate smooth and coherent intermediate frames, performing temporal super-resolution and filling in missing observations.}
    \label{fig:any_res}
\end{figure}

\subsection{ODEWorld for Video Generation}

\textbf{Bidirectional video generation. }
As demonstrated in Fig.~\ref{fig:for_back}, ODEWorld exhibits unique bidirectional flexibility inherent to its ODE-governed architecture. By simply reversing the integration sign ($v_\tau \to -v_\tau$), 
the model can synthesize physically plausible backward video sequences from the same learned latent velocity field. This symmetry validates that ODEWorld internalizes the underlying continuous physics
, rather than merely memorizing forward predictive patterns typical of those discrete predictive architectures.

\textbf{Any-resolution video generation. }
As illustrated in Fig.~\ref{fig:any_res}, ODEWorld leverages its continuous, ODE-governed nature to achieve arbitrary temporal resolution in video generation. Unlike discrete models constrained by fixed step sizes, our model can generate motion at any speed (Fig.~\ref{fig:any_res}(a)) and effectively interpolate missing frames (Fig.~\ref{fig:any_res}(b)). As shown in Fig.~\ref{fig:any_res}(b), even when trained on a down-sampled dataset (e.g., $1/3$ of the original frame rate), ODEWorld successfully recovers the intermediate temporal dynamics, a capability inherently absent in discrete-time baselines.

\textbf{Quantitative comparisons.}
Tab.~\ref{tab:video_baseline} provides a comprehensive evaluation of video prediction quality and efficiency. ODEWorld consistently outperforms baselines in both pixel-level metrics PSNR~\citep{hore2010image} and perception-level metrics LPIPS~\citep{zhang2018unreasonable} across short and long horizons. ODEWorld maintains high structural fidelity and perceptual realism even in long-horizon predictions (@64 frames). Notably, our model achieves substantially lower prediction latency (0.072s for @64 frames) on a single A100 GPU. These results demonstrate that ODEWorld provides accurate long-horizon predictions while maintaining an efficient architecture.

\subsection{ODEWorld for Policy Learning}
We assess how ODEWorld can empower downstream policy learning by inferring latent subgoals as guidance. We instantiate three guidance paradigms: (1) velocity-conditioned $\pi(a|z, c, v)$, (2) single-subgoal $\pi(a|z, c, \hat{z}_{\tau}), \tau=0.25$, and (3) sequential-subgoal $\pi(a|z, c, \{\hat{z}_{\tau_i}\}_{i=1}^n), n=5, \tau_i=0.05i$, where subgoals are generated via ODE integration. $c$ denotes image goal and language instruction.

\textbf{Results on LIBERO-LONG.}
As shown in Tab.~\ref{tab:libero_long_exp}, all ODEWorld variants outperform the baselines on LIBERO-LONG. Among them, the sequential-subgoal paradigm achieves the best performance with an average success rate of 83.6\%. This performance gain demonstrates the high-fidelity temporal consistency of ODEWorld rollouts; by maintaining precise trajectory alignment over horizon, the model provides dense and reliable guidance that does not drift from the task objective. Such consistent sequential subgoal guidance effectively benefits downstream policy learning.

\textbf{Results on real-world robot experiments.}
To further validate the effectiveness of ODEWorld beyond simulation benchmarks, we conduct real-world control experiments on a bi-manual AgileX robot. We use X-VLA~\citep{XVLA} as the policy backbone and augment it with ODEWorld-generated latent subgoals for guidance. We evaluate four manipulation tasks, including: \textit{Shrimp-to-pot}, \textit{Mouse packing}, \textit{Pen insertion}, \textit{Bi-manual rearrangement}. Detailed task settings are provided in App.~\ref{app:realworld_detail}. Using the same policy backbone and training protocol, ODEWorld improves the average success rate from 55\% to 80\% across these four tasks, demonstrating the effectiveness of ODEWorld-generated dynamics guidance in real-world manipulation scenarios.

\section{Conclusions}\label{conclusion}
We introduce ODEWorld, a novel continuous latent world model grounded in physical time. The core of ODEWorld is a new predictive paradigm called PT-Flow, which learns a latent ODE velocity field defined on physical time. It recasts future prediction as temporal integration with the latent velocity via an ODE solver, which shares some similarities to flow matching, but operates in physical time rather than the noise space. Through comprehensive experiments on both simulation and real-world benchmark datasets, we show that ODEWorld enjoys a number of desirable properties, such as interpretable latent velocity fields, prediction with arbitrary directions or temporal resolutions, high-quality reconstruction after long-term prediction, and planning-centric representations that facilitate downstream policy learning. Many of the above properties are fundamentally absent in most discrete-time predictive models. We anticipate that PT-Flow and ODEWorld will inspire a new generation of predictive models, paving the way for a more faithful understanding and modeling of the continuous physical world.

\section*{Acknowledgment}
This work was supported by Wuxi Research Institute of Applied Technologies at Tsinghua University under the Grant 20242001120, and Xiongan AI Institute.


\bibliography{mylib}
\bibliographystyle{unsrt}

\newpage

\section*{NeurIPS Paper Checklist}

\begin{enumerate}

\item {\bf Claims}
    \item[] Question: Do the main claims made in the abstract and introduction accurately reflect the paper's contributions and scope?
    \item[] Answer: \answerYes{} 
    \item[] Justification: We introduce PT-Flow and ODEWorld that are our main contributions.

\item {\bf Limitations}
    \item[] Question: Does the paper discuss the limitations of the work performed by the authors?
    \item[] Answer: \answerYes{} 
    \item[] Justification: Appendix \ref{app:limit}

\item {\bf Theory assumptions and proofs}
    \item[] Question: For each theoretical result, does the paper provide the full set of assumptions and a complete (and correct) proof?
    \item[] Answer: \answerNA{} 
    \item[] Justification: No proofs and assumptions, but we mathematically and rigorously formulate our method in Section \ref{sec:pt_flow}.

    \item {\bf Experimental result reproducibility}
    \item[] Question: Does the paper fully disclose all the information needed to reproduce the main experimental results of the paper to the extent that it affects the main claims and/or conclusions of the paper (regardless of whether the code and data are provided or not)?
    \item[] Answer: \answerYes{} 
    \item[] Justification: In Section \ref{exp} and Appendix \ref{app:exp_detail}.

\item {\bf Open access to data and code}
    \item[] Question: Does the paper provide open access to the data and code, with sufficient instructions to faithfully reproduce the main experimental results, as described in supplemental material?
    \item[] Answer: \answerYes{} 
    \item[] Justification: LIBERO and Agibot World datasets can be open accessed. Our code will be released after acceptance.

\item {\bf Experimental setting/details}
    \item[] Question: Does the paper specify all the training and test details (e.g., data splits, hyperparameters, how they were chosen, type of optimizer) necessary to understand the results?
    \item[] Answer: \answerYes{} 
    \item[] Justification: In Section \ref{exp} and Appendix \ref{app:exp_detail}.

\item {\bf Experiment statistical significance}
    \item[] Question: Does the paper report error bars suitably and correctly defined or other appropriate information about the statistical significance of the experiments?
    \item[] Answer: \answerYes{} 
    \item[] Justification: We report the number of random seeds in experiments.

\item {\bf Experiments compute resources}
    \item[] Question: For each experiment, does the paper provide sufficient information on the computer resources (type of compute workers, memory, time of execution) needed to reproduce the experiments?
    \item[] Answer: \answerYes{} 
    \item[] Justification: We train on 8 A800 GPUs and evaluate inference time on 1 A100 GPU. Details are revealed in Appendix.
    
\item {\bf Code of ethics}
    \item[] Question: Does the research conducted in the paper conform, in every respect, with the NeurIPS Code of Ethics \url{https://neurips.cc/public/EthicsGuidelines}?
    \item[] Answer: \answerYes{} 
    \item[] Justification: Yes, we keep this submission a totally anonymous one.

\item {\bf Broader impacts}
    \item[] Question: Does the paper discuss both potential positive societal impacts and negative societal impacts of the work performed?
    \item[] Answer: \answerYes{} 
    \item[] Justification: We talk about it in Section \ref{conclusion}.
    
\item {\bf Safeguards}
    \item[] Question: Does the paper describe safeguards that have been put in place for responsible release of data or models that have a high risk for misuse (e.g., pre-trained language models, image generators, or scraped datasets)?
    \item[] Answer: \answerNA{} 
    \item[] Justification: The paper poses no such risks.

\item {\bf Licenses for existing assets}
    \item[] Question: Are the creators or original owners of assets (e.g., code, data, models), used in the paper, properly credited and are the license and terms of use explicitly mentioned and properly respected?
    \item[] Answer: \answerYes{} 
    \item[] Justification: We cite them and properly describe them.

\item {\bf New assets}
    \item[] Question: Are new assets introduced in the paper well documented and is the documentation provided alongside the assets?
    \item[] Answer:  \answerNA{} 
    \item[] Justification: The paper does not release new assets

\item {\bf Crowdsourcing and research with human subjects}
    \item[] Question: For crowdsourcing experiments and research with human subjects, does the paper include the full text of instructions given to participants and screenshots, if applicable, as well as details about compensation (if any)? 
    \item[] Answer: \answerNA{} 
    \item[] Justification: The paper does not involve crowdsourcing nor research with human subjects.

\item {\bf Institutional review board (IRB) approvals or equivalent for research with human subjects}
    \item[] Question: Does the paper describe potential risks incurred by study participants, whether such risks were disclosed to the subjects, and whether Institutional Review Board (IRB) approvals (or an equivalent approval/review based on the requirements of your country or institution) were obtained?
    \item[] Answer: \answerNA{} 
    \item[] Justification: The paper does not involve crowdsourcing nor research with human subjects.

\item {\bf Declaration of LLM usage}
    \item[] Question: Does the paper describe the usage of LLMs if it is an important, original, or non-standard component of the core methods in this research? Note that if the LLM is used only for writing, editing, or formatting purposes and does \emph{not} impact the core methodology, scientific rigor, or originality of the research, declaration is not required.
    \item[] Answer: \answerNA{} 
    \item[] Justification: The core method development in this research does not involve LLMs as any important, original, or non-standard components.

\end{enumerate}

\newpage
\appendix
\section*{Appendix}
\section{Experiment Details}\label{app:exp_detail}
\subsection{Datasets}\label{app:dataset_detail}

\paragraph{LIBERO~\citep{liu2023libero}.}
LIBERO is a large-scale language-conditioned robotic manipulation benchmark comprising 130 tasks and 6,500 high-quality human teleoperation demonstrations. These tasks span a diverse set of atomic and compositional skills, including grasping, lifting, placing, opening, and multi-step object rearrangement. The benchmark is organized into five suites, namely LIBERO-Spatial, LIBERO-Object, LIBERO-Goal, LIBERO-90, and LIBERO-Long, each targeting distinct generalization capabilities such as spatial reasoning, object-level variation, goal specification, and long-horizon compositional manipulation. Following the standard evaluation protocol, we use the official LIBERO dataset and re-render all observations at a resolution of $256 \times 256$ to ensure a controlled comparison across methods. We train ODEWorld and all image-generation baselines on the full dataset.

\paragraph{AgiBot-World~\citep{bu2025agibot}. }
For AgiBot-World, we train ODEWorld on the AgiBotWorldChallenge-2025 subset, a large-scale embodied manipulation benchmark built on the AgiBot World Colosseo platform~\citep{bu2025agibot}. The subset used in our experiments contains approximately 30K manipulation trajectories, with an average length of around 260 frames per trajectory. It captures diverse robot-object interaction patterns and provides visual observations across a wide range of real-world manipulation tasks. Compared with existing short-horizon robot datasets, its relatively long trajectories and rich interaction dynamics make it well suited for continuous-time world modeling and future visual prediction.

\paragraph{LIBERO-LONG.}
As a subset of LIBERO, LIBERO-LONG focuses on long-horizon compositional manipulation, where each episode requires completing multiple sequential subgoals involving interactions with several objects. The benchmark consists of ten tasks: (1) Soup and sauce to basket, (2) Box and butter to basket, (3) Turn on the stove and place the pot, (4) Bowl to drawer, (5) Mugs on plates, (6) Book to rack, (7) Mug and pudding, (8) Soup and box to basket, (9) Both pots on the stove, and (10) Mug to microwave.

\subsection{Real-World Settings}
\label{app:realworld_detail}
\begin{figure}[H]
    \centering
    \includegraphics[width=\textwidth]{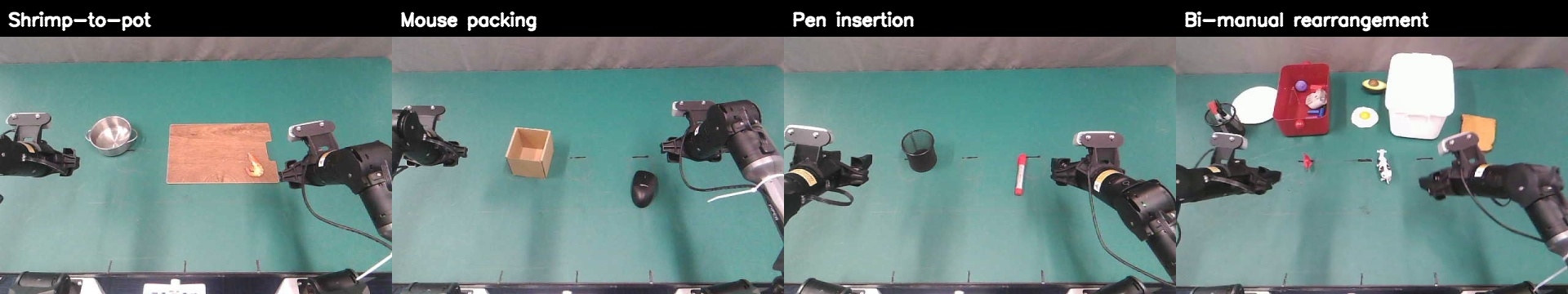}
    \caption{Initial states of the four real-world manipulation tasks. From left to right: \textit{Shrimp-to-pot}, \textit{Mouse packing}, \textit{Pen insertion}, and \textit{Bi-manual rearrangement}.}
    \label{fig:realworld_tasks}
\end{figure}

We follow recent SOTA practice~\citep{black2024zeroshot,zheng2025universal,fengdemystifying} to conduct real-world robotics experiments on a bi-manual AgileX robot equipped with two 6-DoF robotic arms, parallel-jaw grippers, and RGB cameras. We use X-VLA~\citep{XVLA} as the policy backbone and incorporate ODEWorld-generated latent subgoals for sequential guidance. All experiments are conducted under the same training and evaluation protocol for a fair comparison with the original X-VLA policy. We evaluate ODEWorld on four manipulation tasks: (1) \textit{Shrimp-to-pot}: Place a shrimp into a pot; (2) \textit{Mouse packing}: Pack a mouse into a box; (3) \textit{Pen insertion}: Grasp a thin pen and insert it into a narrow container, requiring fine-grained perception and precise control; (4) \textit{Bi-manual rearrangement}: Simultaneously move objects using both arms in a complex scene, evaluating coordinated bi-manual control and multi-object dynamics modeling. Fig.~\ref{fig:realworld_tasks} presents the initial states of these four real-world manipulation tasks.

\subsection{Video Generation Baselines}\label{app:video_gen_baseline}

\paragraph{V-JEPA 2~\citep{assran2025v}.}
We adopt the pre-trained V-JEPA 2 ViT-L/16 checkpoint as the visual representation backbone and fine-tune it on LIBERO demonstration videos from the agentview camera. The model follows the standard online-target encoder design of V-JEPA 2, where the target encoder is updated by EMA. During training, we use short video clips cropped to $256 \times 256$ and apply a first-tubelet-visible masking strategy, forcing the predictor to infer future latent tokens from the initial visible spatio-temporal context. To incorporate task information, we extract language features with a frozen CLIP ViT-B/32 text encoder~\citep{radford2021learning} and inject them into the predictor through zero-initialized cross-attention layers. Since V-JEPA 2 predicts latent representations rather than pixels, we additionally train a lightweight convolutional decoder for RGB reconstruction. The encoder-predictor stack is frozen when training this decoder. To improve robustness to errors in predicted latents, we perturb the training latents with Gaussian noise. The decoder is optimized with an L1 reconstruction loss over video frames and an LPIPS~\citep{zhang2018unreasonable} loss on temporally subsampled frames.

\paragraph{LDP~\citep{xie2025latent}.}
We implement LDP based on the official codebase and make several modifications for fair comparison under high-resolution LIBERO video modeling. Specifically, we increase the visual input resolution to $256\times256$ and enlarge the visual token budget in the planner/IDM pipeline accordingly. To handle the high-dimensional SD-VAE latent representation, we introduce a spatial convolutional projection before policy learning, which compresses the latent feature map and stabilizes training. We also add explicit goal-image conditioning by encoding the goal frame with the same VAE and injecting the goal latent into the planning module. During inference, LDP performs open-loop rollout, where future latent trajectories are autoregressively predicted without intermediate environment feedback. These settings are used consistently for all reported LDP results.

\subsection{Policy Learning Baselines}\label{app:policy_baseline}

\paragraph{GLCBC.}
GLCBC is a goal-conditioned language behavioral cloning baseline implemented within the LBP policy architecture but without the high-level planner. The current image observation is encoded by a ResNet-34~\citep{he2016deep}, while the language instruction is encoded by CLIP~\citep{radford2021learning}. In addition, we use a predefined final goal image as goal conditioning and encode it with the DecisionNCE image encoder~\citep{li2024decisionnce}. The language and goal-image embeddings are concatenated and injected into the policy through a FiLM~\citep{perez2018film} module to modulate the visual features. Finally, the low-level diffusion policy takes the conditioned visual semantics and current proprioception as inputs, and predicts the diffusion noise for action generation.

\paragraph{SuSIE~\citep{black2024zeroshot}.}
We utilize the official codebase with minimal modifications, altering only the datasets. For simulation experiments, we fine-tune the released SuSIE checkpoint on the LIBERO dataset. 

\paragraph{Seer~\citep{tian2025predictive}.}
We use Seer as a vision-language-action baseline following its original implementation. The model predicts future visual representations from the current observation and language instruction, and then uses an inverse-dynamics head to predict actions conditioned on both the current observation and the predicted future features.

\paragraph{VPP~\citep{VPP}.} 
VPP learns an implicit inverse dynamics model conditioned on predicted future representations inside VDMs. To predict a more precise future, VPP fine-tunes the pre-trained video foundation model on robot datasets along with internet human manipulation data. We follow the official codebase to finetune it on the LIBERO dataset.

\subsection{Computation Resource}
We train models on 8 A100 GPUs and evaluate infernece time in Tab.~\ref{tab:video_baseline} on 1 A100 GPU.

\begin{figure}[H]
    \centering
    \includegraphics[width=\textwidth]{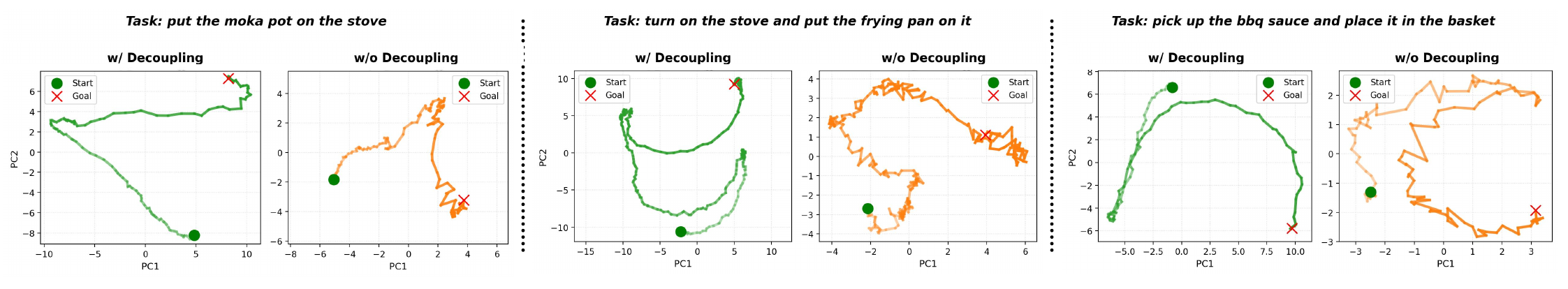}
    \caption{Latent trajectory visualization of ablation on dynamical representation decoupling.}
    \label{fig:ablation}
\end{figure}

\begin{table}[H]
\centering
\caption{Ablations on dynamical representation decoupling and first-order supervision}
\label{tab:ablation_decouple}
\resizebox{\textwidth}{!}{
\begin{tabular}{x{180}x{45}x{45}x{45}x{45}x{45}x{45}}
\toprule
\multirow{2}{*}{Method}
& \multicolumn{3}{c}{Short Fragment @16frames}
& \multicolumn{3}{c}{Long Fragment @64frames} \\
\cmidrule(lr){2-4} \cmidrule(lr){5-7}
& PSNR$\uparrow$ & LPIPS$\downarrow$ & FPS$\uparrow$ & PSNR$\uparrow$ & LPIPS$\downarrow$ & FPS$\uparrow$ \\
\midrule
w/o dynamical representation decoupling & \textbf{20.65} & 0.127 & 7.301 & \textbf{19.53} & 0.212 & 2.290 \\
w/o first-order supervision & 12.71 & 0.751 & \textbf{33.67} & - & - & - \\
\rowcolor{gray!20} ODEWorld (Ours) & 20.53 & \textbf{0.109} & \textbf{33.67} & 19.46 & \textbf{0.134} & \textbf{13.83} \\
\bottomrule
\end{tabular}
}   
\end{table}

\section{Ablation Studies}\label{app:ablation}
\begin{table}[t]
\centering
\caption{Comparisons on the choices of vision feature encoder.}
\label{tab:ablation_encoder}
\begin{tabular}{x{75}x{45}x{45}x{45}x{45}}
\toprule
\multirow{2}{*}{Encoder}
& \multicolumn{2}{c}{Short Fragment @16frames}
& \multicolumn{2}{c}{Long Fragment @64frames} \\
\cmidrule(lr){2-3} \cmidrule(lr){4-5}
& PSNR$\uparrow$ & LPIPS$\downarrow$ & PSNR$\uparrow$ & LPIPS$\downarrow$ \\
\midrule
LIV & 19.90 & 0.158 & 18.25 & 0.237 \\
SigLIP 2 & 20.38 & 0.133 & 19.21 & 0.168 \\
\rowcolor{gray!20} DINOv2 (Ours) & \textbf{20.53} & \textbf{0.109} & \textbf{19.46} & \textbf{0.134} \\
\bottomrule
\end{tabular}
\end{table}

\begin{table}[t]
\centering
\caption{Ablation on the number of latent tokens.}
\label{tab:ablation_token}
\begin{tabular}{x{75}x{45}x{45}x{45}x{45}}
\toprule
\multirow{2}{*}{\# Tokens}
& \multicolumn{2}{c}{Short Fragment @16frames}
& \multicolumn{2}{c}{Long Fragment @64frames} \\
\cmidrule(lr){2-3} \cmidrule(lr){4-5}
& PSNR$\uparrow$ & LPIPS$\downarrow$ & PSNR$\uparrow$ & LPIPS$\downarrow$ \\
\midrule
4 & \textbf{20.99} & 0.114 & \textbf{19.68} & 0.148 \\
\rowcolor{gray!20} 1 (Ours) & 20.53 & \textbf{0.109} & 19.46 & \textbf{0.134} \\

\bottomrule
\end{tabular}
\end{table}

\begin{figure}[t]
    \centering
    \includegraphics[width=\textwidth]{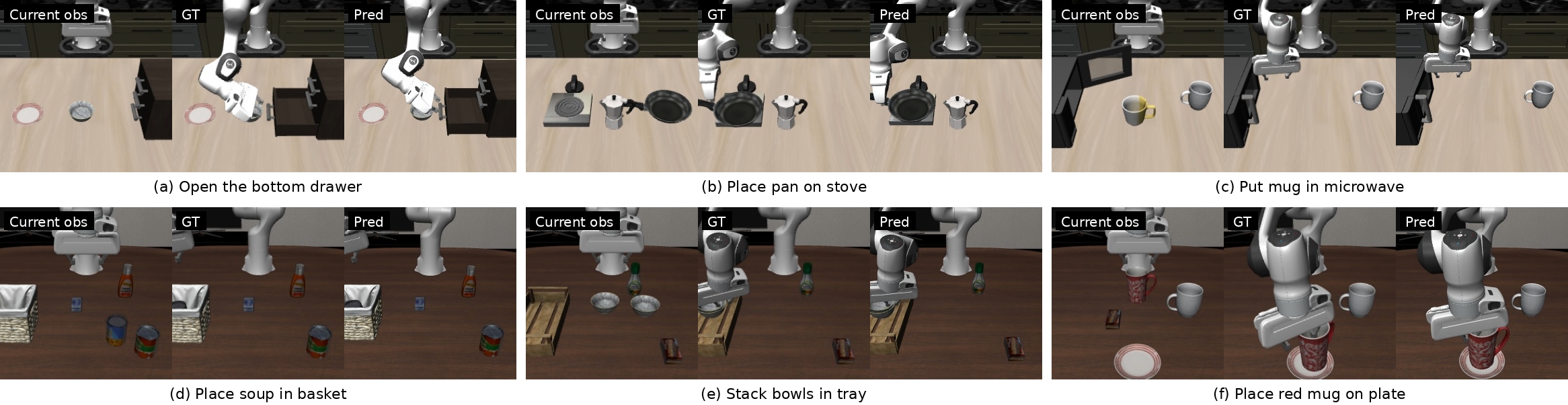}
    \caption{\small Qualitative results of the goal image predictor. Given the current observation and a language instruction, the goal image predictor generates a semantically consistent goal image for ODEWorld.}
    \label{fig:goal_predictor}
\end{figure}

\begin{table}[t]
\centering
\caption{Comparison between ground-truth image goals and predicted language goals.}
\label{tab:ablation_goal}
\begin{tabular}{x{95}x{45}x{45}x{45}x{45}}
\toprule
\multirow{2}{*}{Goal Conditioning}
& \multicolumn{2}{c}{Short Fragment @16frames}
& \multicolumn{2}{c}{Long Fragment @64frames} \\
\cmidrule(lr){2-3} \cmidrule(lr){4-5}
& PSNR$\uparrow$ & LPIPS$\downarrow$ & PSNR$\uparrow$ & LPIPS$\downarrow$ \\
\midrule
Predicted Goal Image & 21.06 & 0.110 & 19.91 & 0.142 \\
GT Goal Image & 20.53 & 0.109 & 19.46 & 0.134 \\
\bottomrule
\end{tabular}
\end{table}

\begin{table}[t]
\centering
\caption{Ablation on the detach in velocity learning.}
\label{tab:ablation_detach}
\begin{tabular}{x{75}x{45}x{45}x{45}x{45}}
\toprule
\multirow{2}{*}{Methods}
& \multicolumn{2}{c}{Short Fragment @16frames}
& \multicolumn{2}{c}{Long Fragment @64frames} \\
\cmidrule(lr){2-3} \cmidrule(lr){4-5}
& PSNR$\uparrow$ & LPIPS$\downarrow$ & PSNR$\uparrow$ & LPIPS$\downarrow$ \\
\midrule
w/o detach & 21.18 & 0.107 & 20.00 & 0.123 \\
w/ detach & 20.53 & 0.109 & 19.46 & 0.134 \\
\bottomrule
\end{tabular}
\end{table}

\subsection{Ablation on Dynamical Representation Decoupling}
To validate the effectiveness of our architectural design, we evaluate the impact of the dynamical representation decoupling module in ODEWorld. We observe that without explicitly isolating the static context $s_0$ from the state space, the learned latent trajectories become significantly non-smooth and jittery as shown in Fig.~\ref{fig:ablation}. 
The primary reason is that not all nuanced visual features in a raw video observation can be seamlessly modeled as an ODE function. 
Raw feature spaces inherently contain a substantial amount of noisy, time-invariant, and non-ODE-compatible information, such as static backgrounds and object textures. 
Forcing a continuous velocity field to fit these static details fundamentally disrupts the dynamics learning process. 
As shown in Tab.~\ref{tab:ablation_decouple}, the proposed decoupling strategy not only significantly improves inference efficiency (significantly higher FPS) but also enhances perceptual consistency (much lower LPIPS scores, especially for long-horizon prediction).
We note that the non-decoupled variant attains marginally higher PSNR, which is expected rather than contradictory: as an MSE-based pixel-level metric, PSNR is dominated by large static regions (e.g., backgrounds and textures) that this variant retains and reconstructs faithfully, whereas its worse LPIPS reflects the temporal and semantic inconsistency that such per-frame fidelity does not capture.
Consequently, without decoupling, the latent structure fails to effectively scaffold the direct first-order supervision to optimize the velocity field. 
By introducing the initial state-conditioned encoder and decoder, our decoupling mechanism ensures that the latent representation focuses exclusively on essential dynamical changes, which is crucial for learning a well-behaved, temporally smooth, and physically meaningful latent flow. 

\subsection{Ablation on First-Order Supervision}
To verify the effectiveness of the proposed first-order supervision, we implement consistency loss below to replace the reconstruction loss $\mathcal{L}_\text{dyn-recon}$ and velocity loss $\mathcal{L}_v$:
\begin{equation}
    \mathcal{L}_\text{consistency}=\mathbb{E}_{s_0,s_t\sim \mathcal{D}, k\in\{1,\cdots,K\}, t} \left\|g_\text{dyn}\left(z_t + \int_{t}^{t+k} v_\theta(z_t, t; z_0, c) dt; s_0\right)-s_{t+k}\right\|^2
\label{eq:consistency_loss}
\end{equation}
where $z_t=f_\text{dyn}(s_t;s_0)$, $z_0=f_\text{dyn}(s_0;s_0)$ and $K=4$. 
Specifically, this consistency-loss variant minimizes the error of reconstructing the subsequent $K$ frames obtained by planning through the ODE latent space, which has been adopted in many works on neural ODE~\citep{chen2018neural,rubanova2019latent}.
In Tab.~\ref{tab:ablation_decouple}, we observe that it suffers from optimization difficulty, yielding poor short-horizon prediction and visually implausible long-horizon rollouts (so we fill those results in Tab.~\ref{tab:ablation_decouple} with ``-'').
We attribute this to two aspects. 
First, the consistency loss couples two distinct objectives — prediction and reconstruction — forcing the model to learn a latent space that is simultaneously reconstructable and ODE-plannable. Second, this coupling is especially harmful in our setting: unlike standard dynamical systems where the goal is to fit one specific governing ODE, ODEWorld only needs to fit some admissible ODE in the latent space, so the target is underdetermined and the entangled objective adds too strict constraints to settle on a consistent solution.
In contrast, ODEWorld decouples these objectives via a reconstruction loss and a direct first-order supervision loss, optimizing each independently, which leads to more stable optimization and, in turn, better video generation quality.

\subsection{Ablations on Vision Encoders}
We examine how ODEWorld is compatible with different choices of vision feature extractors by replacing DINOv2 with SigLIP 2~\citep{tschannen2025siglip} and LIV~\citep{ma2023liv}. As shown in Tab.~\ref{tab:ablation_encoder}, ODEWorld maintains robust video generation quality across all three encoders. While the gaps are small, DINOv2 achieves the best overall performance on both short- and long-horizon prediction, consistent with its widely recognized effectiveness as a visual encoder.

\subsection{Ablation on Token Numbers}
We increase the number of tokens from 1 to 4 and keep other configurations unchanged. Results in Tab.~\ref{tab:ablation_token} show comparable performance between 1- and 4-token models, indicating that a \textit{single} latent token is already sufficient to achieve strong performance across the evaluated scenarios. This also indicates that our key designs make ODEWorld capable of efficiently capturing the underlying dynamics from pixels with very compact latent representation.

\subsection{Ablation on Goal Descriptions}\label{app:lang-goal}
ODEWorld naturally supports image goals, where the target image provides explicit visual guidance for future dynamics prediction. However, in practical scenarios, goals are often specified by language instructions. To extend ODEWorld to language-based goal descriptions, we introduce a lightweight goal image predictor to bridge the language and visual modalities.

Specifically, since ODEWorld adopts DINOv2 as the visual backbone rather than a large-scale vision-language model, we avoid introducing additional multimodal alignment complexity into the dynamics model. Instead, we train a lightweight goal image predictor consisting of four cross-attention layers, which takes the current observation and language instruction as inputs and predicts the corresponding goal image. As shown in Fig.~\ref{fig:goal_predictor}, the goal image predictor generates semantically meaningful goal images from the current observation and language instruction, providing effective visual guidance for ODEWorld.

We further evaluate ODEWorld using the predicted goal images for goal conditioning and compare it with the standard setting using ground-truth goal images. As reported in Tab.~\ref{tab:ablation_goal}, the two settings achieve nearly identical video prediction performance on both short- and long-horizon rollouts. These results demonstrate that the proposed goal image predictor effectively bridges language and visual goal representations, enabling ODEWorld to naturally support language-specified goals without sacrificing prediction quality.


\subsection{ODEWorld without Detach in Velocity Learning}\label{app:detach}
In PT-Flow, we use the stop-gradient operation $\texttt{sg}(\cdot)$ on the target latent velocity to stabilize the optimization, following common practices in representation learning~\citep{chen2021exploring}. Interestingly, we find that PT-Flow remains effective even without stop-gradient, as shown in Tab.~\ref{tab:ablation_detach}. This suggests that the proposed objective itself can provide effective constraints for learning meaningful representations, as the first-order velocity constraint directly regularizes the temporal evolution in latent space.

\section{Analysis of Representation Collapse}
\label{app:collapse}

Following RankMe~\citep{Rankme}, we use the effective rank of mean-centered representations to assess potential representation collapse. We randomly sample 5,000 representations across planning horizons $h\in\{1,4,8,16,32\}$. As shown in Tab.~\ref{tab:rankme_analysis}, ODEWorld's 768-dimensional dynamics latent $z$ consistently maintains an effective rank above 400 across all horizons. In Tab.~\ref{tab:rankme_compare}, we further compare ODEWorld with V-JEPA 2 representations and DINOv2 CLS features under the same sampling protocol. ODEWorld achieves a substantially higher effective rank than V-JEPA 2 and DINOv2, indicating that its learned latent dynamics representations remain diverse and expressive with a lower risk of representation collapse.

\section{Additional Results}
\begin{figure}[t]
    \centering
    \includegraphics[width=\textwidth]{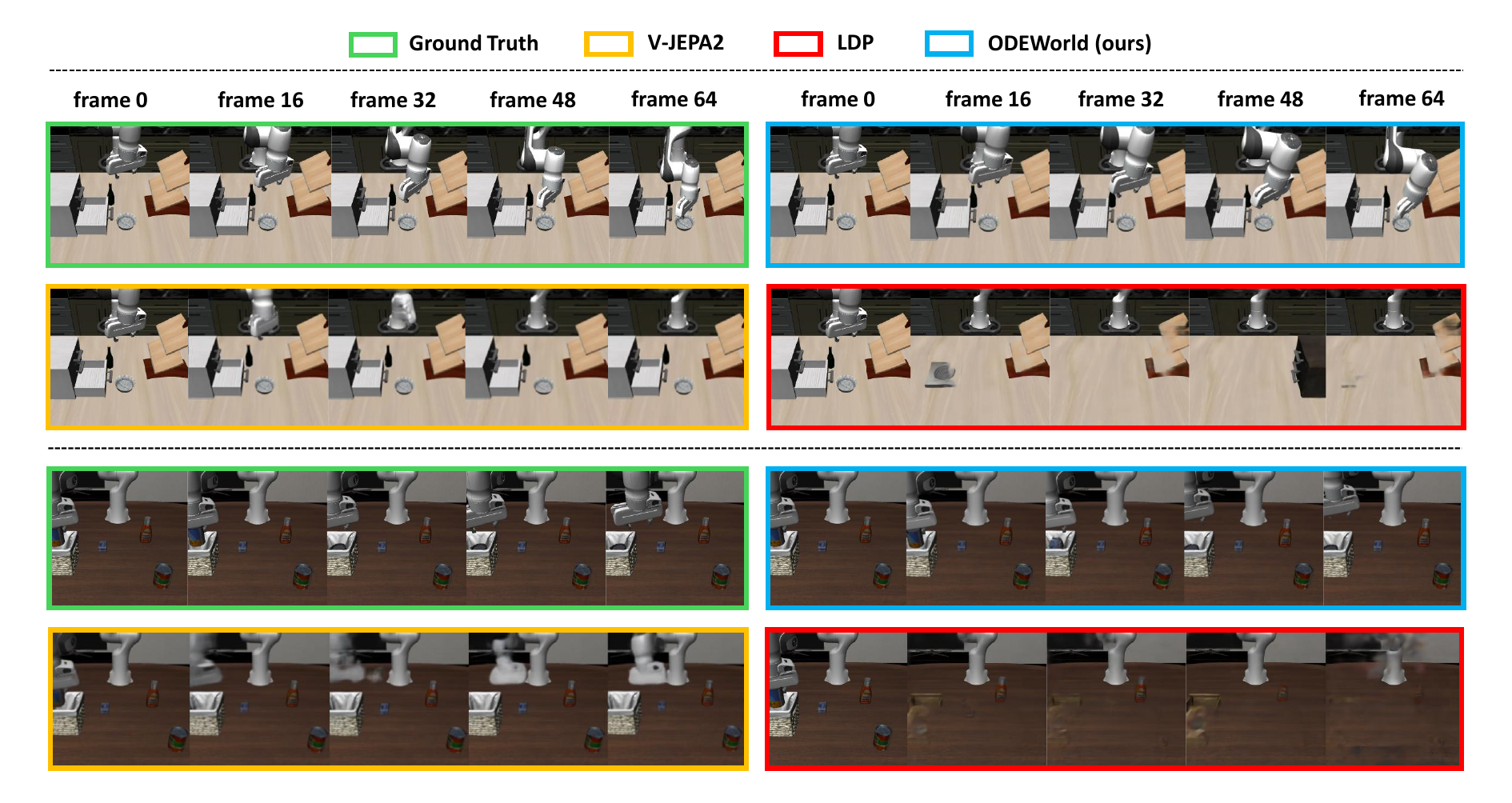}
    \caption{\small ODEWorld enables long-horizon video generation in the LIBERO simulation environment, producing temporally coherent and physically plausible videos that closely match the ground-truth observations. In contrast, our baseline methods LDP and V-JEPA 2 often suffer from blurry visual predictions and dataset-induced hallucinations in long-horizon generation, especially at later frames, such as frame 64.}
    \label{fig:app_libero_video_gen}
\end{figure}

\begin{figure}[!htbp]
    \centering
    \includegraphics[width=0.85\textwidth]{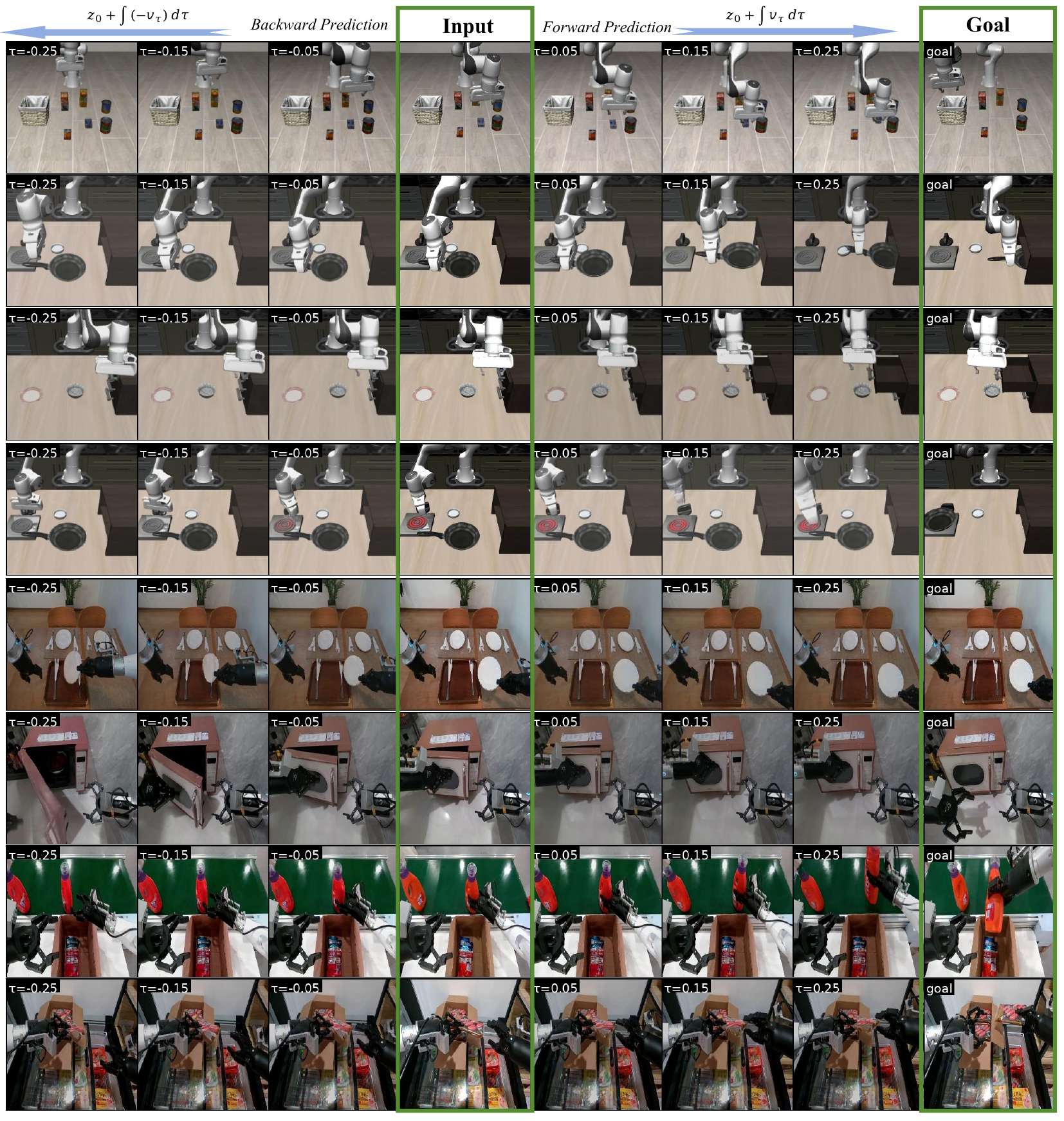}
    \caption{Additional visualization results of bidirectional prediction.}
    \label{fig:for_back_all}
\end{figure}
\subsection{Additional Video Generation Quality Comparisons}\label{app:video_gen_qual}
In Fig.~\ref{fig:app_libero_video_gen}, we provide additional comparisons for long-horizon video generation in the LIBERO simulation environment. These visualizations complement the main paper and further illustrate that ODEWorld can produce temporally coherent and physically plausible rollouts, while baseline methods often suffer from visual degradation and temporal inconsistency over extended horizons.

\begin{table}[t]
\centering
\caption{Centered RankMe effective rank analysis of ODEWorld dynamics latents across different planning horizons.}
\label{tab:rankme_analysis}
\begin{tabular}{x{100}x{45}x{45}x{45}x{45}x{45}}
\toprule
Horizon $h$ 
& 1 & 4 & 8 & 16 & 32 \\
\midrule
Effective Rank / Dim.
& 417.2/768
& 413.9/768
& 422.1/768
& 433.1/768
& 439.5/768 \\
\bottomrule
\end{tabular}
\end{table}

\begin{table}[!t]
\centering
\caption{Centered RankMe effective rank comparison with V-JEPA 2 and DINOv2 representations.}
\label{tab:rankme_compare}
\begin{tabular}{c|c|c}
\toprule
Representation & Dimension & Effective Rank (Mean) \\
\midrule
V-JEPA 2 & 1024 & 203.7 \\
DINOv2 CLS & 768 & 376.1 \\
ODEWorld $z$ & 768 & 425.2 \\
\bottomrule
\end{tabular}
\vspace{-5pt}
\end{table}

\subsection{Additional Bidirectional Video Generation Results}\label{app:video_any_res_gen}
In Fig.~\ref{fig:for_back_all}, we show additional bidirectional video generation results on both LIBERO and AgiBot World.
We perform closed-loop planning that updates the initial state every integration step $\tau=0.05$. When we plan backward, we integrate velocity $-v_\theta(z_0, 0; z_0, c)$ for every integration step and reset the planned latent state as $z_0$ for the next integration step.

\section{Limitations}\label{app:limit}
Although our proposed framework provides a promising architecture for continuous-time predictive modeling, the current version of ODEWorld still has several limitations and can be further enhanced in future research. First, while the proposed ODEWorld demonstrates some impressive capabilities and efficiency, its scale and amount of training data (two robotics datasets) remain small as compared to many large video-based world models. This limitation stems primarily from computational resource constraints. Future studies can be conducted to further scale the model size and train on larger and more diverse datasets. Second, our current version of ODEWorld uses a frozen DINOv2~\citep{oquab2024dinov2} image encoder to encode raw observations. As DINO is designed to encode single-frame images without much consideration for temporal consistency, it could produce noisy and temporally inconsistent embeddings for sequential modeling. In our study, we introduce the Savitzky–Golay derivative filter~\citep{saramaki1993finite} to enhance the smoothness and approximation quality of the ground truth velocity target to mitigate the deficiencies of the DINO encoder. Further investigations can consider replacing DINO with an image encoder with temporal consistency design~\citep{li2024decisionnce, li2025robo} or jointly tuning the DINO encoder during model training.
Lastly, the current version of ODEWorld does not incorporate action conditioning and primarily focuses on video generation. Future studies can consider injecting action conditioning into a pre-trained ODEWorld model, which can further unlock its capability for control tasks.

\end{document}